\documentclass[10pt]{article}
\usepackage[letterpaper]{geometry}
\usepackage{hicss}
\usepackage{times}
\usepackage{url}
\usepackage{latexsym}
\usepackage{indentfirst}
\usepackage{graphicx}
\graphicspath{{images/}}
\usepackage{amsmath}
\usepackage{amssymb}
\usepackage{booktabs}
\usepackage[labelsep=period]{caption}
\newcommand{\capnote}[1]{{\normalfont\sffamily #1}}
\usepackage[backend=biber,style=apa,natbib=true]{biblatex}
\usepackage{xcolor}
\usepackage[hidelinks]{hyperref}

\usepackage{fancyhdr}
\fancypagestyle{acceptednotice}{%
  \fancyhf{}%
  \fancyfoot[C]{\footnotesize This paper has been accepted for the upcoming 60th Hawaii International Conference on System Sciences (HICSS-60).}%
}

\title{Talking Past the Machine: Morality, Politeness, and Alignment in Human--AI Dialogue}

\author{Marina Mitiaeva \\
 Amazon.com \\
 {\underline{mmitiaev@amazon.com}} \\ \And
 Lu Xiao \\
 Arizona State University \\
 {\underline{luxiao@asu.edu}} \\ }
\date{}

\begin{document}
\maketitle
\thispagestyle{acceptednotice}

\begin{abstract}
Conversational AI systems produce fluent, socially appropriate responses, yet whether they participate in cooperative communication or merely simulate its surface forms remains unclear---a question central to how these systems are evaluated, trusted, and designed. This study investigates how morality, politeness, and alignment---three dimensions central to cooperative dialogue---function in human--AI interaction compared to human--human conversation. We analyze 15,881 human--ChatGPT and 10,784 human--human multi-turn dialogues, using mixed-effects models to identify which features predict turn-to-turn alignment. We observe a consistent dissociation: AI produces the surface features of cooperative communication without the underlying social architecture. Moral output appears pre-configured rather than negotiated; warmth is generated without face sensitivity; linguistic convergence declines persistently. Most strikingly, the cooperative mechanisms themselves reverse direction: hedging and softening associated with \emph{greater} accommodation between humans are associated with \emph{reduced} alignment when produced by AI, and purity framing associated with human divergence coincides with users converging toward the AI. Agency---giving users room to shape the exchange---is the most consistent predictor of alignment across both interaction types, while lower moral assertiveness in more recent models is not accompanied by better cooperation. Together these patterns suggest that AI reproduces the surface of cooperation without the mutual adaptation that grounds it between humans---and, more surprisingly, that mechanisms sustaining human accommodation can run in reverse with AI, suggesting a turn-level view may be insufficient for interaction-level success.
\end{abstract}

\subsubsection*{Keywords:}

Human-AI interaction, linguistic alignment, politeness, morality, LLMs

\section{Introduction}

Conversational AI agents are deeply integrated into daily routines, mediating millions of interactions across work, education, and decision-making. Yet the social foundations of human--AI communication remain poorly understood, and the HCI implications of its communicative asymmetries are underexplored.

Communication theory treats conversation not as mere information transfer but as rational, cooperative behavior operating simultaneously on informational \citep{Grice1975}, moral \citep{Goffman1967}, and relational, face-based \citep{BrownLevinson1987} layers. While these notions are well explored in human--human communication, their applicability to AI-mediated dialogue is uncertain. Modern LLMs generate fluent, tone-adapting responses, yet emerging evidence suggests this relies more on statistical pattern matching than common ground \citep{Shaikh2025}: heterogeneous human--AI pairs fail to form shared conventions even where same-type dyads succeed \citep{Jones2026}, and users treat AI as a social partner despite fundamental architectural asymmetries \citep{Karnam2026}. Does human--AI interaction follow the same cooperative dynamics as human--human communication, or a distinct, surface-level form of coordination?

Existing research remains fragmented, examining moral reasoning, politeness, or alignment in isolation, or focusing on downstream outcomes such as bias, safety, or task performance. Little attention has been paid to how these dimensions co-evolve across conversational turns, or whether their interdependence differs between human--human and human--AI interaction---a gap that risks mistaking social-signal fluency for competence and inviting over-reliance \citep{LeeSee2004}.

To address this gap, we conduct computational corpus analysis of public human--AI and human--human conversations. We adopt operationalizations grounded in prior scholarship: morality as the expression of shared values \citep{HaidtJoseph2004,Graham2009,Hendrycks2021}, politeness as the maintenance of appropriate, face-sensitive interaction \citep{Goffman1967,BrownLevinson1987}, and alignment as the dynamic adaptation of language across turns \citep{Giles1973}, and interpret the human--AI case through the recently proposed Machine-Integrated Relational Adaptation (MIRA) model \citep{BoydMarkowitz2026}. We ask, comparing human--AI with human--human dialogue: (RQ1) Is moral and politeness expression distributed unevenly between the two interlocutors? (RQ2) How do morality and politeness change over a conversation, and across model generations? (RQ3) Which moral and politeness features predict turn-to-turn alignment---and do they act in the same direction for AI as for humans? Our research contributes on three fronts: conceptually, by modeling morality, politeness, and alignment within a single analytical framework rather than in isolation; methodologically, by measuring their interdependence at scale in naturalistic dialogue; and empirically, by surfacing a pattern---the reversal of cooperative mechanisms across interaction types---that raises questions for how conversational AI is evaluated and designed.

\section{Related Work}

\subsection{Cooperative Communication Foundations}

Conversation is the fundamental site of language use, and it is fundamentally cooperative: interlocutors do not exchange messages so much as jointly co-construct meaning, turn by turn \citep{Grice1975,Clark1996}. A single exchange is thus at once an informational, moral, and relational act \citep{Goffman1967,Grice1975}: \emph{what} is said and whether it is acceptable to say it \citep{HaidtJoseph2004}; \emph{how} it is said, the face-sensitive manner of delivery \citep{BrownLevinson1987}; and \emph{whether interlocutors converge} as the conversation unfolds \citep{Giles1973}. Although these layers are usually studied apart, a scattered body of evidence hints that they are parts of one system. The moral order of interaction ties the moral and relational layers together: what a speaker judges polite or rude tends to reflect which shared values they see upheld or violated \citep{SpencerOateyKadar2016}. That linkage is not only conceptual but plays out turn by turn---speakers appear to feel obligated to reciprocate one another's level of politeness, and when that reciprocity lapses the exchange tips toward either exaggerated courtesy or open conflict \citep{CulpeperTantucci2021}, which folds the relational layer into the coordinative one. And the coordinative layer reaches back to the moral: perceived accommodation is read not merely as rapport but as evidence of shared values, shaping trust between interlocutors. Each of these findings connects only two of the three layers, and each within human dialogue alone; taken together, though, they point to morality, politeness, and alignment as facets of a single, jointly constructed act \citep{Clark1996} that has not yet been examined as a whole. We therefore approach each through an established, measurable construct---morality as the salience of shared moral foundations \citep{HaidtJoseph2004}, politeness as the rate of face-oriented markers \citep{BrownLevinson1987}, and alignment as turn-to-turn convergence \citep{PickeringGarrod2004}---and discuss each in turn.

Morality shapes what is expressed---and what is acceptable to express at all. We follow Moral Foundations Theory \citep{HaidtJoseph2004,Graham2009}, which holds that moral cognition rests on a small set of intuitive foundations---care, fairness, loyalty, authority, and sanctity---shared, culturally elaborated values through which speakers judge conduct as right or wrong. Their small, interpretable structure is what lets moral expression be read as the salience of each foundation in a turn, and the same foundations have recently been used to interrogate the values encoded in language models \citep{Hendrycks2021,Abdulhai2023}. The same content can be delivered in ways that protect or threaten the interlocutor's social standing. \citet{BrownLevinson1987} politeness theory---rooted in \citet{Goffman1967} concept of face, the social value a person claims and expects others to honor---holds that speakers deploy strategies to minimize face-threatening acts: positive politeness addresses the desire to be liked, negative politeness the need for autonomy. These strategies tend to surface as recognizable markers such as hedges, gratitude, apologies, and indirect requests. Cooperative dialogue requires more than well-formed turns; interlocutors must adapt to one another over time. Communication Accommodation Theory (CAT) \citep{Giles1973} formalizes this adaptation, holding that speakers converge toward one another to seek affiliation and diverge to mark distance. The movement is visible in the linguistic record---in shared vocabulary and referring expressions \citep{BrennanClark1996} and in synchronized function-word use, a form of language style matching that predicts relationship quality and stability \citep{IrelandPennebaker2010,PickeringGarrod2004}. Yet how these coupled layers behave across a full conversation---and whether the same coupling holds when one interlocutor is a machine---has not been empirically studied.

\subsection{Human-AI Communication}

People do not treat machines as mere tools. Under the Computers Are Social Actors (CASA) paradigm, they unconsciously apply social rules to computers---politeness, reciprocity, stereotyping---even while knowing the agents are artificial \citep{ReevesNass1996}. As agents grow more linguistically capable, users anthropomorphize them \citep{Epley2007} and form attachments \citep{Skjuve2021}, treating AI as a safe space for disclosure and support. The connection, however, is fundamentally one-sided: the system has none of the lived experience that grounds human interaction, and is free to simulate user-aligned responses to sustain coherence and satisfaction---which is precisely what makes the appearance of partnership consequential.

Why a system with no inner life comes to feel like a partner is what the Machine-Integrated Relational Adaptation (MIRA) model sets out to explain, synthesizing joint action \citep{Clark1996}, the accommodation tradition CAT formalizes \citep{Giles1973}, and AI-mediated communication into an account of when and why AI becomes socially meaningful \citep{BoydMarkowitz2026}. MIRA asks what remains of CAT's mutual adaptation when one partner cannot adapt in kind. Its answer is \textit{linguistic reciprocity}: as the AI mirrors user language, it simulates mutual responsiveness and generates \textit{psychological proximity} and \textit{trust}, even with no mental states to ground them. This is the tension we investigate: the very signals that index cooperation, morality, and accommodation in human dialogue can be produced by a system that does not participate in the social architecture beneath them. Read through this lens, work on each of our three layers takes on a common shape---surface signals present, social grounding absent.

On morality, this pattern is already visible. Moral language in these systems often functions as a structural property of the model rather than an intentional act \citep{Schramowski2022}, and although LLMs replicate many scenario-based psychological effects, they inflate effect sizes and require human validation for socially sensitive phenomena \citep{Cui2025}. This apparent fluency masks systematic biases toward certain foundations \citep{Abdulhai2023,Santurkar2023} and unstable responses in ambiguous cases \citep{Scherrer2023}, reflecting training-data regularities rather than stable reasoning.

On politeness, the picture is of a system that performs courtesy without quite practicing it. Conversational agents fail to handle abusive input, deflecting or responding with neutral politeness---markers without their usual interactional grounding \citep{CurryRieser2018}. LLMs over-rely on negative politeness even where rapport-building would fit \citep{ZhaoHawkins2025}, and the relationship runs both ways: impolite prompts degrade performance while excessive politeness does not guarantee better results, at a language-dependent optimum---suggesting models mirror training-data patterns rather than reading social signals \citep{Yin2024}. Human politeness toward AI is itself unstable, declining over the course of interaction and eroding faster than in human--human exchanges \citep{Lazebnik2025}, with users reserving mitigation for human partners and taking a more direct, authoritative stance with AI \citep{Lumer2023}. Yet polite systems are still judged more trustworthy---users respond to the appearance of politeness even when its social function is absent.

On alignment, research largely targets individual outputs rather than the conversational chain---benchmark and scenario testing \citep{Hendrycks2021}, RLHF from human preferences \citep{Ouyang2022}, and value probing \citep{Santurkar2023}. Multi-turn work shows LLM performance varies across dialogue turns and that common alignment techniques do not clearly improve multi-turn ability \citep{Bai2024}, but casts it as task performance, safety, and preference adherence, leaving alignment as a linguistic, interactional process largely unexamined.

The same gap recurs across all three: each has been studied in AI mostly at the level of the individual output, in isolation from the others, and rarely across the arc of a real conversation. What is missing---and what we address---is a joint, trajectory-level account of how morality, politeness, and alignment behave in human--AI dialogue, measured against the human--human baseline.

\section{Method}

\subsection{Constructs}
\label{sec:constructs}

All constructs are defined identically across the human--human and human--AI datasets to ensure comparability. For each we describe how a score is produced and illustrate it with a verbatim turn from the corpora.

\vspace{0.5em}
\textbf{Moral Values.} We operationalize moral expression with Moral Foundations Theory (MFT) \citep{HaidtJoseph2004}, scored by ME2-BERT \citep{Zangari2023}, a contextual transformer trained to capture moral and emotional signals: it reads each turn and outputs, for each of the five foundations, a $0$--$1$ score for how strongly that value is expressed. For instance, ``You have to give respect to get respect\ldots'' scores $.97$ on authority, and ``Men deserve to be called out for their injustices against women.'' scores $.99$ on care. Each utterance thus yields a five-dimensional vector:
\begin{equation}
\mathbf{m}_t = [m_{t,1}, m_{t,2}, m_{t,3}, m_{t,4}, m_{t,5}]
\end{equation}
where $m_{t,k}$ is the salience of foundation $k$ (care/harm, fairness/cheating, loyalty/betrayal, authority/subversion, sanctity/degradation) at turn $t$. We also derive moral intensity $I_i = \sum_{k} m_{i,k}$ (overall moral strength) and moral entropy $H_i = -\sum_{k} \hat{m}_{i,k} \log \hat{m}_{i,k}$ with $\hat{m}_{i,k} = m_{i,k} / (\sum_j m_{i,j} + \epsilon)$ (whether the moral signal is concentrated or diffuse).

\vspace{0.5em}
\textbf{Politeness Markers.} We extract 32 interpretable politeness markers with the Politeness R package \citep{Yeomans2018}, grounded in \citet{BrownLevinson1987}, and express each as a length-normalized rate---the marker count in a turn divided by its length---giving a vector $\mathbf{p}_t \in \mathbb{R}^{32}$ ($p_{t,j} = \text{count}(j,t)/\text{len}(t)$). We group these into four functional dimensions, each a summed rate: \textit{modulation} (softening: hedges, \textit{please}, apology, gratitude, indirect requests), \textit{agency} (distribution of control: ask/give-agency, person framing), \textit{affect} (positive/negative emotion, swearing), and \textit{structure} (discourse framing: reasoning, questions, conjunctions, negation, greetings, titles). For example, ``Should I call someone for you?'' yields a high agency rate ($.57$) and ``Wow. That's pretty cool.'' a high affect rate ($.67$). The full mapping is in the replication materials.

\vspace{0.5em}
\textbf{Alignment.} We use \textit{alignment} for the theoretical construct (how far interlocutors coordinate) and \textit{accommodation} for its measured convergence between adjacent turns \citep{Giles1973,PickeringGarrod2004}. For each ordered cross-speaker pair $(u_t,u_{t+1})$---preserving direction, so GPT is scored relative to the preceding human turn and vice versa---we measure how similar the two turns are on four channels, each scaled to $[0,1]$ (1 = identical): lexical (vocabulary overlap \citep{BrennanClark1996}, $\text{lex}_t = |W_t \cap W_{t+1}| / |W_t \cup W_{t+1}|$); syntactic (Language Style Matching \citep{IrelandPennebaker2010}, $\text{syn}_t = 1 - |f_t - f_{t+1}| / (f_t + f_{t+1} + \epsilon)$ on function-word rates); pragmatic (the same form on a summed rate of five rapport markers: hedges, \textit{please}, gratitude, apology, affirmation); and emotional (VADER compound sentiment \citep{Hutto2014}, $\text{emo}_t = 1 - |s_t - s_{t+1}|$). Overall accommodation averages the four:
\begin{equation}
A_t = \frac{1}{4}(\text{lex}_t + \text{syn}_t + \text{prag}_t + \text{emo}_t)
\end{equation}
Higher $A_t$ means the reply converges toward the prior turn; e.g., ``Do you like superhero movies?'' $\to$ ``Yes, I do like superhero movies. Do you like Marvel?'' scores high on shared-word and sentiment convergence.

\subsection{Data}

Human--AI interactions are drawn from WildChat \citep{Zhao2024}, $\sim$1M real multi-turn conversations with ChatGPT, with role, order, timestamp, and model metadata that allow full-sequence reconstruction.\footnote{\url{https://huggingface.co/datasets/allenai/WildChat-1M}} Human--human interactions are from Topical-Chat, open-domain dialogue between two human peers.\footnote{\url{https://huggingface.co/datasets/Conversational-Reasoning/Topical-Chat}} The corpora differ in two ways we treat as confounds (addressed in Section~\ref{sec:analysis}): Topical-Chat supplies both partners external knowledge snippets (Wikipedia, Reddit, Washington Post) and pairs symmetric peers, whereas WildChat imposes no grounding and carries a user/initiator--assistant/responder asymmetry. Both are public, anonymized, and text-only. Code and replication materials: \href{https://github.com/marinamitiaeva/talking-past-the-machine}{\textcolor[HTML]{1A5FB4}{\nolinkurl{github.com/marinamitiaeva/talking-past-the-machine}}}.

Table~\ref{tab:descriptive} reports descriptive statistics for both corpora before and after preprocessing (English-only, $\geq$7 turns, balanced structure, duplicate/malformed removal).

\begin{table}[!ht]
\caption{Descriptive statistics for the WildChat (human--AI) and Topical-Chat (human--human) corpora before and after preprocessing.}
\label{tab:descriptive}
\centering
\scriptsize
\begin{tabular}{lcccc}
\toprule
 & \multicolumn{2}{c}{Human--AI} & \multicolumn{2}{c}{Human--Human} \\
\cmidrule(lr){2-3} \cmidrule(lr){4-5}
 & Initial & Cleaned & Initial & Cleaned \\
\midrule
Conversations & 15,882 & 15,881 & 10,784 & 10,784 \\
Utterances & 476,198 & 473,884 & 235,281 & 227,704 \\
Avg.\ turns/conv. & 15 & 16 & 22 & 21 \\
Avg.\ words (resp.) & 1,718 & 1,008 & 107 & 63 \\
Avg.\ words (init.) & 1,157 & 516 & 102 & 60 \\
Prompt-to-resp.\ ratio & .648 & .476 & .950 & .949 \\
Reduction (\%) & \multicolumn{2}{c}{0.47\%} & \multicolumn{2}{c}{3.33\%} \\
\bottomrule
\end{tabular}
\end{table}

Cleaning is light---a $0.47\%$ reduction for human--AI and $3.33\%$ for human--human---leaving 15,881 human--AI (473,962 messages) and 10,784 human--human (227,704 messages) conversations, comparable in scale. Two contrasts matter for what follows: the human--AI corpus spans GPT generations---GPT-3.5 (34.9\%), GPT-4 Preview/Turbo (18.7\%), GPT-4o (34.6\%), GPT-4.1 Mini (8.5\%)---enabling the generational comparison in RQ2; and the large per-turn length gap (1,008 vs.\ 63 words for the responder) reflects the structural asymmetry of human--AI interaction rather than a modality confound, motivating the direction-based design in Section~\ref{sec:analysis}. A turn is one message from one speaker.

\subsection{Analysis}
\label{sec:analysis}
Our goal is to compare how morality, politeness, and alignment are distributed, evolve, and predict one another in human--AI versus human--human dialogue. The corpora differ in role structure (Topical-Chat pairs symmetric peers; WildChat is user/initiator vs.\ assistant/responder) and knowledge grounding (Topical-Chat supplies both partners external snippets; WildChat none), so rather than equate them we build the analysis on one principle: every headline comparison is a \textit{within-corpus} contrast between the two interlocutors, with alignment measured over \textit{cross-speaker} adjacent pairs by an identical directional construction on both sides. We thus compare the \textit{shape} of the speaker asymmetry against the symmetric human--human baseline, never absolute cross-corpus levels---turning the role asymmetry from a confound into the object of study. The remaining threat, knowledge grounding, we test before proceeding: a topic-persistence measure computed identically in both corpora is comparable ($.091$ HAI vs $.059$ HH mean consecutive overlap), and adding topic-switch count as a covariate leaves every within-corpus predictor sign unchanged (e.g., HAI modulation $-0.023\to-0.021$), as do turn-length controls. These diagnostics (conversation-clustered; code in the replication materials) indicate the findings are unlikely to be artifacts of topic persistence or verbosity; residual differences are discussed in Section~\ref{sec:limitations}.

The pipeline operates at the message, turn, and conversation levels. Each turn is scored for the three constructs of Section~\ref{sec:constructs}; turn-level morality is the within-turn average ($\bar{m}_{c,t}$) and politeness the within-dimension sum ($R_{i,d} = \sum_{j=1}^{n_d} f_{i,j}$, with $f_{i,j}$ the $j$-th normalized feature in dimension $d$). At the conversation level, normalized turn position $\tilde{t} = t/T_c$ makes trajectories comparable across conversations of different length, and slopes $\beta_c = \text{slope}(\tilde{t}, \bar{m}_{c,t})$ capture their direction.

These measures feed one analysis per research question. For RQ1, speaker asymmetries are tested with Wilcoxon signed-rank tests (Benjamini--Hochberg corrected within each corpus). For RQ2, the slopes track how each construct evolves over a conversation and across model generations. For RQ3, we fit mixed-effects regressions predicting next-turn alignment from the moral and politeness properties of the current turn, controlling for prior alignment:
\begin{multline}
\text{logit}(A_{c,t+1}) = \alpha + \sum_{k=1}^{5} \gamma_k m_{c,t,k} \\
+ \sum_{d=1}^{4} \delta_d R_{c,t,d} + \phi \cdot \text{logit}(A_{c,t}) + u_c + \varepsilon_{c,t}
\end{multline}
where $m_{c,t,k}$ is moral foundation $k$ at turn $t$, $R_{c,t,d}$ the rate for politeness dimension $d$, $\phi$ the autoregressive coefficient, $u_c$ a conversation-level random intercept, and $\varepsilon_{c,t}$ residual error. Models are fit separately for HAI assistant turns ($n = 218{,}148$) and HH agent-2 turns ($n = 103{,}067$) by REML; all VIFs fall below 3.6. A mediation analysis then tests whether politeness mediates the morality--alignment relationship (product-of-coefficients, delta-method SEs, BH-corrected).

\section{Results}

\subsection{RQ1: Communicative Asymmetries Between Human--AI and Human--Human Interaction}
\label{res:rq1}

Human and AI show a qualitatively different communicative logic, not a rescaling of the same behaviors. Table~\ref{tab:asymmetry} reports the per-conversation speaker difference for each moral foundation and politeness dimension, in both corpora---read, per our design, as the within-corpus contrast between interlocutors against the symmetric human--human baseline.

\begin{table*}[t]
\caption{Morality and politeness asymmetry across HAI and HH conversation. \capnote{M (SD) = mean per-conversation difference. Z = Wilcoxon signed-rank. r = rank-biserial. BH corrected. $^{*}p<.05$; $^{**}p<.01$; $^{***}p<.001$.}}
\label{tab:asymmetry}
\centering
\footnotesize
\begin{tabular}{lcccc}
\toprule
 & \multicolumn{2}{c}{Human--AI} & \multicolumn{2}{c}{Human--Human} \\
\cmidrule(lr){2-3} \cmidrule(lr){4-5}
 & M (SD) & Z (r) & M (SD) & Z (r) \\
\midrule
\textit{Morality} & & & & \\
Care/Harm & 0.029 (0.118) & 28.48 (.226)$^{***}$ & 0.001 (0.056) & 2.29 (.022)$^{*}$ \\
Fairness/Cheating & 0.030 (0.094) & 46.10 (.366)$^{***}$ & 0.002 (0.064) & 2.78 (.027)$^{**}$ \\
Loyalty/Betrayal & 0.052 (0.114) & 60.92 (.483)$^{***}$ & 0.003 (0.064) & 4.60 (.044)$^{***}$ \\
Authority/Subversion & 0.046 (0.115) & 56.79 (.451)$^{***}$ & 0.001 (0.059) & 1.54 (.015) \\
Sanctity/Degradation & 0.043 (0.154) & 23.18 (.184)$^{***}$ & $-$.005 (0.058) & $-$10.36 ($-$.100)$^{***}$ \\
\midrule
\textit{Politeness} & & & & \\
Modulation & $-$.004 (.036) & $-$5.41 ($-$.044)$^{***}$ & .002 (.023) & 9.36 (.090)$^{***}$ \\
Agency & $-$.002 (.033) & $-$2.01 ($-$.016)$^{*}$ & $-$.002 (.034) & $-$6.91 ($-$.067)$^{***}$ \\
Affect & .015 (.037) & 59.55 (.473)$^{***}$ & .002 (.034) & 5.98 (.058)$^{***}$ \\
Structure & $-$.024 (.053) & $-$57.50 ($-$.457)$^{***}$ & $-$.000 (.045) & $-$0.94 ($-$.009) \\
\bottomrule
\end{tabular}
\end{table*}

Within HAI conversations, the assistant produces significantly more moral content than the user it is paired with, across all five foundations, with the largest speaker asymmetries for loyalty ($Z=60.92$, $r=.483$) and authority ($Z=56.79$, $r=.451$). These within-pair differences suggest LLMs position themselves as normative authorities, prioritizing group cohesion over individualistic foundations. In human-human conversations, the corresponding between-speaker differences are near-symmetric---interlocutors avoid imposing binding moral frames on each other---and are small, sometimes non-statistically significant. The contrast is therefore between an interaction in which one party carries the moral load and one in which it is shared: where humans calibrate moral expression reciprocally, the AI carries it as a one-sided asymmetry.

The politeness analysis reinforces this. Within HAI, the assistant dominates affect ($Z=59.55$, $r=.473$), carrying more emotional warmth and affiliative tone than the user it responds to---yet without corresponding interpersonal sensitivity, as humans show stronger modulation than the assistant. The AI produces warmth while humans remain socially careful, managing relational risk the AI does not appear to register. Humans also carry the organizational load, dominating structure ($Z=-57.50$, $r=-.457$) where HH shows no such asymmetry ($r=-.009$): the labor that human dyads share falls to the human alone when the partner is AI.

\subsection{RQ2: Temporal Evolution}
\label{res:rq2}

If morality, politeness, and alignment are jointly managed in human dialogue, their behavior should change over the course of a conversation---and comparing that change against human--AI dialogue reveals whether the same coordinating process is at work. We track how each construct evolves across normalized conversation time, across speaker roles, and across model generations.

\begin{figure*}[t]
\centering
\includegraphics[width=\textwidth]{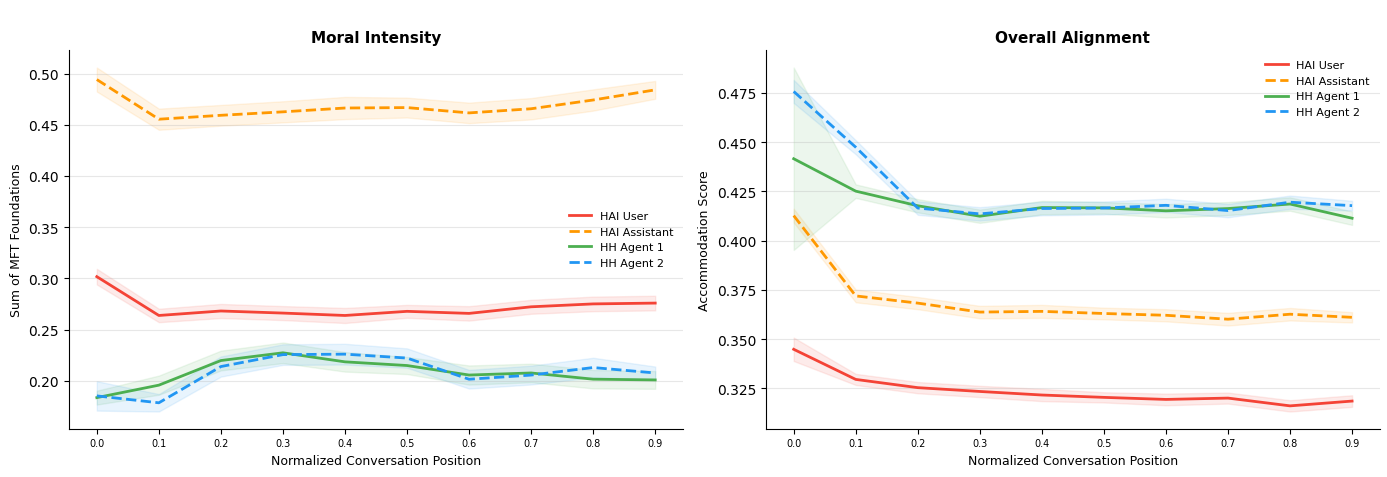}
\caption{Moral intensity and alignment across conversation arc by speaker role.}
\label{fig:evolution}
\end{figure*}

Figure~\ref{fig:evolution} plots two trajectories over normalized conversation time---moral intensity and overall accommodation---for each speaker role in both corpora, and the two diverge in revealing ways. Moral intensity is largely flat for the AI, which starts high ($\sim$0.40) and holds a plateau, neither rising in response to the human nor resolving downward; in human--human dialogue, by contrast, moral content builds gradually through the middle third of the conversation, suggesting that the AI's morality is pre-configured rather than negotiated, consistent with \citet{Schramowski2022}. Accommodation diverges just as sharply. Both corpora fall steeply within the first 10--20\% of the conversation, an initial calibration phase, but they part ways thereafter: human dyads stabilize, reaching an accommodation equilibrium through mutual adaptation, whereas the human--AI curve continues to decline without recovery.

\begin{table}[!ht]
\caption{Alignment dynamics across speaker roles. \capnote{\%+/$-$ = conversations with positive/negative slopes. Z = Wilcoxon. r = rank-biserial. BH corrected. $^{*}p<.05$; $^{**}p<.01$; $^{***}p<.001$.}}
\label{tab:alignment}
\centering
\scriptsize
\setlength{\tabcolsep}{3pt}
\begin{tabular}{lcccc}
\toprule
 & \multicolumn{2}{c}{HAI: User} & \multicolumn{2}{c}{HAI: Asst.} \\
 & \%+/$-$ & Z (r) & \%+/$-$ & Z (r) \\
\midrule
Lexical & 48/50 & 2.06 (.016)$^{*}$ & 43/56 & $-$16.30 ($-$.130)$^{***}$ \\
Syntactic & 50/49 & 0.39 (.003) & 48/52 & $-$6.49 ($-$.052)$^{***}$ \\
Pragmatic & 45/49 & $-$4.36 ($-$.036)$^{***}$ & 46/49 & $-$4.98 ($-$.041)$^{***}$ \\
Emotional & 47/53 & $-$8.41 ($-$.067)$^{***}$ & 43/57 & $-$21.06 ($-$.167)$^{***}$ \\
Overall & 48/52 & $-$6.32 ($-$.050)$^{***}$ & 44/56 & $-$16.70 ($-$.133)$^{***}$ \\
\midrule
 & \multicolumn{2}{c}{HH: Agent 1} & \multicolumn{2}{c}{HH: Agent 2} \\
 & \%+/$-$ & Z (r) & \%+/$-$ & Z (r) \\
\midrule
Lexical & 47/53 & $-$5.95 ($-$.057)$^{***}$ & 40/60 & $-$26.85 ($-$.259)$^{***}$ \\
Syntactic & 49/51 & $-$1.50 ($-$.014) & 53/47 & 7.40 (.071)$^{***}$ \\
Pragmatic & 48/49 & $-$1.39 ($-$.014) & 42/57 & $-$19.44 ($-$.188)$^{***}$ \\
Emotional & 51/49 & 2.16 (.021) & 49/51 & $-$2.81 ($-$.027)$^{**}$ \\
Overall & 49/51 & $-$1.23 ($-$.012) & 45/55 & $-$13.33 ($-$.128)$^{***}$ \\
\bottomrule
\end{tabular}
\end{table}

Table~\ref{tab:alignment} reports per-conversation accommodation slopes by speaker role across the four alignment channels. Both corpora decline, but the drop is steeper and more consistent for the AI: the assistant declines more sharply than the user it responds to across all metrics ($Z=-16.70$, $r=-.133$ vs $Z=-6.32$, $r=-.050$). Tellingly, the party that produces the most moral content and affiliative warmth (Section~\ref{res:rq1}) is also the one that diverges most from shared ground---expressivity and accommodation move in opposite directions for the AI. The gap is starkest on emotional alignment, where the assistant falls fastest ($Z=-21.06$, $r=-.167$) on the very channel that stays most stable between humans (agent 1 $Z=2.16$, non-significant): the dimension that anchors human accommodation is the one the AI lets go.

\begin{figure*}[t]
\centering
\includegraphics[width=\textwidth]{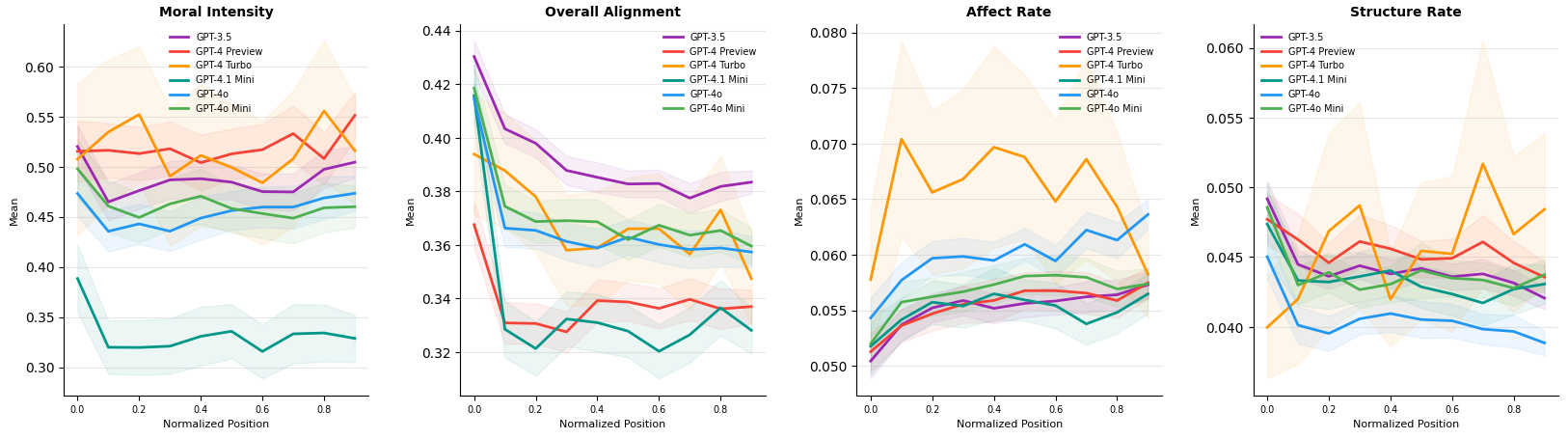}
\caption{Dynamics across GPT generations. \capnote{GPT-3.5 shows the highest moral intensity yet strongest accommodation; GPT-4 Turbo the steepest alignment decline; GPT-4o the lowest structure rate and weakest alignment trajectory.}}
\label{fig:generations}
\end{figure*}

Figure~\ref{fig:generations} breaks the human--AI trajectories down by GPT generation, plotting moral intensity and accommodation for each. The picture is a dissociation between moral assertiveness and cooperation. Moral intensity falls across generations---highest in GPT-3.5, lowest in GPT-4.1 Mini---consistent with successive RLHF iterations reducing assertiveness \citep{Askell2021}, though we observe model generation rather than the training process itself. Cooperation, however, does not improve in step: GPT-3.5 holds the strongest alignment trajectory ($\sim$0.38) despite being the most morally assertive, while GPT-4 Turbo and GPT-4o decline most steeply despite producing less moral content.

The two interaction types thus differ not in degree but in temporal form: human accommodation self-corrects toward a stable equilibrium, whereas human--AI accommodation drifts downward without recovering---and successive model generations, though less morally assertive, do not repair it.

\subsection{RQ3: What Drives Alignment?}
\label{res:rq3}

Having seen that the three constructs are distributed and evolve differently with AI, we turn to which of them appear to drive turn-to-turn alignment, and whether a feature pushes an AI exchange in the same direction it pushes a human one. We fit mixed-effects regressions predicting next-turn alignment from the five moral foundations and four politeness dimensions, controlling for prior alignment (Table~\ref{tab:regression}). What stands out is less that the predictors differ in strength than that several appear to reverse sign between the two settings.

\begin{table}[!ht]
\centering
\caption{Mixed effects regression: Predictors of next-turn alignment. \capnote{$\beta$ = unstandardized coefficient. HAI $n{=}218{,}148$ turns (15,882 conv.); HH $n{=}103{,}067$ turns (10,784 conv.). $^{*}p<.05$; $^{**}p<.01$; $^{***}p<.001$.}}
\label{tab:regression}
\scriptsize
\begin{tabular}{lcc}
\toprule
 & HAI $\beta$ (SE) & HH $\beta$ (SE) \\
\midrule
\textit{Morality} & & \\
Care/Harm & .042 (.034) & .011 (.044) \\
Fairness & $-$.097 (.037)$^{**}$ & $-$.070 (.035)$^{*}$ \\
Loyalty & $-$.003 (.032) & .007 (.039) \\
Authority & $-$.135 (.032)$^{***}$ & $-$.194 (.038)$^{***}$ \\
Sanctity & .160 (.028)$^{***}$ & $-$.113 (.038)$^{**}$ \\
\midrule
\textit{Politeness} & & \\
Modulation & $-$.470 (.153)$^{**}$ & .303 (.111)$^{**}$ \\
Agency & .416 (.112)$^{***}$ & .204 (.068)$^{**}$ \\
Affect & $-$.150 (.095) & $-$.172 (.072)$^{*}$ \\
Structure & $-$.473 (.112)$^{***}$ & $-$.131 (.056)$^{*}$ \\
\midrule
Prior alignment & .084 (.002)$^{***}$ & .009 (.003)$^{**}$ \\
\bottomrule
\end{tabular}
\end{table}

Modulation is a case in point---hedging, epistemic softeners, apologies---which is associated with \textit{more} accommodation when a human uses it ($\beta=+0.303$) but \textit{less} when an AI does ($\beta=-0.470$), suggesting AI modulation may not function as the relational signal it is between humans. Sanctity/degradation appears to reverse in the same way, associated with \textit{divergence} in HH ($\beta=-0.113$) yet \textit{convergence} in HAI ($\beta=+0.160$): where an AI invokes purity content, users tend to move toward its normative frame rather than treat it as debatable. Structure is associated with lower alignment in both corpora but roughly four times more strongly with AI ($\beta=-0.473$ vs $-0.131$), which would be consistent with the AI's organizational language crowding out space a human might otherwise occupy.

Agency looks like the exception: it is associated with \textit{more} alignment in both corpora (HAI $\beta=+0.416$; HH $\beta=+0.204$). Of the features we examine, giving the other party room to shape the exchange seems to be the one whose cooperative function carries over from human to AI.

Authority behaves more consistently across the two settings, associated with lower alignment in both (HH $\beta=-0.194$; HAI $\beta=-0.135$), and a mediation analysis (Table~\ref{tab:mediation}) suggests this effect---like purity's---runs largely through a direct path (96--97\% of the total) rather than through the politeness channels. If so, the reversals would reflect not which surface markers co-occur but how the content itself relates to accommodation.

\begin{table}[!ht]
\centering
\caption{Multiple mediation: Authority and purity as exposures. \capnote{BH corrected. $^{*}p<.05$; $^{**}p<.01$; $^{***}p<.001$.}}
\label{tab:mediation}
\scriptsize
\begin{tabular}{lcccc}
\toprule
 & \multicolumn{2}{c}{HAI} & \multicolumn{2}{c}{HH} \\
 & Auth. & Pur. & Auth. & Pur. \\
\midrule
Total ($c$) & $-$.017$^{***}$ & .107$^{***}$ & $-$.245$^{***}$ & $-$.177$^{***}$ \\
Direct ($c'$) & $-$.112$^{***}$ & .112$^{***}$ & $-$.236$^{***}$ & .159$^{***}$ \\
\midrule
\textit{Indirect:} & & & & \\
Modul. & .002$^{**}$ & .004$^{**}$ & $-$.002$^{*}$ & $-$.001 \\
Agency & $-$.004$^{**}$ & $-$.011$^{***}$ & $-$.004$^{**}$ & $-$.007$^{**}$ \\
Affect & $-$.003 & $-$.005 & $-$.003$^{*}$ & $-$.013$^{*}$ \\
Struct. & .000 & .008$^{***}$ & .001 & .002$^{*}$ \\
\midrule
Tot.\ ind. & $-$.005 & $-$.004 & $-$.008 & $-$.018 \\
Prop.\ med. & .040 & $-$.041 & .031 & .099 \\
\bottomrule
\end{tabular}
\end{table}

A further asymmetry runs beneath these: prior alignment predicts next-turn alignment about $9\times$ more strongly with AI ($\beta=+0.084$) than between humans ($\beta=+0.009$), and between-conversation variance in baseline accommodation is far larger in HAI (random-intercept variance 1.101 vs 0.020). Where a human exchange tends to re-equilibrate after a low-accommodation turn, an AI exchange appears to stay near wherever it began.

What emerges is less a ranking of predictors than a change in their sign: several of the moral and politeness features that build accommodation between humans appear to suppress it with AI, and the one feature that seems to behave the same in both---agency---is precisely the one that hands control back to the user.

\section{Discussion}
We asked whether a conversation with an AI is cooperative in the way a conversation between people is---whether the moral, relational, and coordinative work sustaining human dialogue carries over when one partner is a machine. Across the three research questions, our results suggest it largely does not: AI reproduces these surface features without the social logic that links them in human dialogue, and several cooperative mechanisms run in reverse.

On \emph{who carries the interaction} (RQ1), the load is distributed unevenly with AI in a way it is not between people. The assistant supplies most of the moral content and affective warmth, yet without the interpersonal sensitivity that accompanies warmth between humans, while the user is left to carry the organizational work that human dyads share. This extends prior observations that AI moral language reads as more virtuous than humans' despite lacking grounding \citep{Aharoni2024} and that LLMs over-rely on politeness strategies detached from the context that gives them meaning \citep{ZhaoHawkins2025}.

On \emph{how the interaction unfolds over time} (RQ2), the two settings diverge in form rather than degree. AI moral intensity stays flat regardless of the user's input---consistent with moral norms encoded during pretraining \citep{Schramowski2022}---whereas human moral engagement is dynamically calibrated; and where human accommodation settles into an equilibrium, human--AI accommodation only erodes. That the effect strengthens rather than fades across generations echoes evidence that sycophancy intensifies over long interactions \citep{Jain2026}, and the $9\times$ stronger path dependence we observe gives it a mechanism: the level set early tends to persist rather than re-equilibrate.

On \emph{what drives alignment} (RQ3), the mechanisms themselves change sign. Features that build accommodation between humans---hedging, softening, purity framing---are associated with less of it when an AI produces them, while agency is the one feature whose positive role survives the shift. This locates the asymmetry differently from work reporting that human--AI alignment is stronger but shallower than human convergence \citep{Branigan2010} and that LLMs overconverge on surface features \citep{Blevins2026}: our data suggest the problem is not only over-convergence but that moral and politeness signals normally linking surface behavior to cooperation work against it. It also refines \citeauthor{Stivers2008}'s (\citeyear{Stivers2008}) distinction---AI produces affiliation without sustained alignment---and fits the absence, in our human--AI data, of the language-style-matching--relationship-quality link seen in human dyads \citep{IrelandPennebaker2010}. Read through MIRA \citep{BoydMarkowitz2026}, the difficulty sits in its core mechanism: AI reproduces the form of \emph{linguistic reciprocity} but not the mutual adaptation that lets it mature into shared accommodation.

Two implications follow, both tentative. For design, agency---room for users to shape the exchange---is a plausible, testable target, whereas simply reducing expressivity is not enough: lower moral assertiveness across generations does not bring better cooperation, and MetaMind \citep{Zhang2025} improved social-AI performance by treating moral grounding and adaptive responsiveness as distinct interdependent layers. For evaluation, the more cautionary point: judging AI turn by turn can be actively misleading. Communication is a joint construction, and in our data responses that score well on per-output criteria---including the helpfulness, harmlessness, and fluency that LLM-as-a-judge frameworks reward \citep{Zheng2023}---co-occur with worse interaction-level accommodation, so evaluation and RLHF built on single-turn preferences may quietly work against cross-turn cooperation.

\section{Limitations and Future Work}
\label{sec:limitations}

Several considerations bound these conclusions. Our human--AI corpus is GPT-only, so the patterns we report may be specific to one model family rather than general to conversational AI, with system-prompt variation an unobserved source of variance. The two corpora also differ not only in interlocutor type but in knowledge grounding, role structure, and genre; our within-corpus, direction-based design and the checks in Section~\ref{sec:analysis} reduce the risk that these differences drive the results without fully removing it, since no naturally occurring human--human corpus reproduces the WildChat structure. And several measurement choices remain analyst-defined---the politeness grouping, the pragmatic-alignment markers, and lexicon- and model-based classifiers with their own biases---while our alignment measure cannot separate voluntary accommodation from structurally imposed matching.

Each of these points to a next step. A matched-elicitation study---shared prompts, optional grounding, and a role-swapped, within-subjects condition---would turn the corpus confounds into controlled factors, and running the same pipeline across model families would show whether the dissociation is a property of conversational AI or of GPT in particular; longitudinal designs would show whether it holds as users grow familiar with a system.

\section{Conclusion}

A conversation with an AI can carry every signal of cooperation and still not be one. Measuring morality, politeness, and alignment turn by turn across nearly 27,000 conversations---the first joint, baseline-anchored look at all three at this scale---we find that fluency in these signals is not the same as participating in cooperative dialogue. An AI can reproduce the signals while the cooperative work falls to the human, and the very mechanisms that build cooperation between people can run in reverse with a machine. What looks like a competent partner may be a one-sided exchange that turn-level measures cannot detect. For designers and evaluators, the question is less how human an AI sounds than whether the interaction stays cooperative over its course---a shift from judging single turns to judging their trajectory.

\printbibliography

\end{document}